\documentclass[a4paper, 10pt, conference]{ieeeconf}      

\IEEEoverridecommandlockouts                              

\usepackage{amsmath} 
\usepackage{amssymb}  
\usepackage{graphicx}
\usepackage{comment}
\usepackage{subfig} 
\usepackage{multicol}
\usepackage{algorithm}
\usepackage{algpseudocode}
\usepackage{soul}
\usepackage{xcolor}

\title{\LARGE \bf
Auction-Consensus Algorithm with Learned Bidding Scheme for Multi-Robot Systems
}

\author{Jose Rodriguez$^1$, Constantine Tarawneh$^2$, Sven Koenig$^3$, Wenjie Dong$^1$, and Qi Lu$^{4,*}$
\thanks{$^1$Department of Electrical and Computer Engineering,
        The University of Texas at Rio Grande Valley (UTRGV), Edinburg, USA
        {\tt\small \{jose.rodriguez53, wenjie.dong\}@utrgv.edu}}%
\thanks{$^2$Department of Mechanical Engineering, UTRGV, Edinburg, TX, USA     {\tt\small constantine.tarawneh@utrgv.edu}}%
\thanks{$^3$Department of Computer Science, Donald Bren School of Information and Computer Sciences, University of California, Irvine
        {\tt\small svenk@uci.edu}}%
\thanks{$^4$Department of Computer Science, UTRGV, Edinburg, TX, USA
        {\tt\small qi.lu@utrgv.edu}}%
\thanks{$^*$Corresponding author}
}

\begin{document}

\maketitle
\thispagestyle{empty}
\pagestyle{empty}

\begin{abstract}
Multi-Robot Task Allocation (MRTA) is a central challenge in decentralized multi-agent systems, where teams of robots must cooperatively assign and execute tasks under limited communication while optimizing global performance objectives. Auction-consensus algorithms, such as the Consensus-Based Bundle Algorithm (CBBA), provide scalable decentralized coordination with provable convergence, but rely on hand-crafted greedy scoring functions that often lead to suboptimal task allocations. This paper proposes a learning-enhanced auction-consensus framework in which CBBA’s deterministic bidding mechanism is replaced by a neural bidding policy trained using reinforcement learning. Under a centralized training and decentralized execution paradigm, agents learn to compute task bids from partial local observations while retaining the standard auction and consensus phases for decentralized coordination. The learned bidding policy is trained using Proximal Policy Optimization with rewards shaped by proximity to globally optimal solutions obtained via mixed-integer linear programming. Multiple neural architectures are evaluated, including a Neural Additive Model, the Long Short-Term Memory (LSTM) model, and the Set Transformer Model. Experimental results across varying swarm sizes demonstrate that learned bidding policies can improve solution quality over classical CBBA while preserving decentralized execution. The proposed approach highlights the effectiveness of integrating reinforcement learning with classical distributed coordination algorithms, offering a scalable pathway toward higher-quality decentralized multi-robot task allocation.

\end{abstract}


\section{Introduction}
\label{Introduction}

Multi-Robot Task Allocation (MRTA) is a fundamental challenge in multi-agent systems, where teams of robots must efficiently divide and execute tasks in a shared environment. MRTA approaches can generally be categorized into centralized and decentralized frameworks. Centralized methods rely on a global coordinator that collects system-wide information and computes an optimal or near-optimal assignment. While often achieving high solution quality, such approaches typically suffer from scalability limitations, increased communication overhead, and vulnerability to single points of failure~\cite{SwarmReview}. In contrast, decentralized strategies distribute decision-making across agents using local information and limited communication, enabling improved robustness, adaptability, and scalability in large-scale robotic teams. As a result, decentralized coordination has become a widely adopted framework in multi-robot systems.

MRTA frameworks typically aim to optimize one of two primary global objectives. The first seeks to minimize the sum of all agents' travel or execution costs, often referred to as the min-sum objective. The second aims to minimize the maximum distance traveled or time required by any single agent, typically known as the min-max objective.

A widely studied and influential decentralized approach for MRTA is the Consensus-Based Bundle Algorithm (CBBA)~\cite{CBBA}, a market-based coordination method that relies on greedy task allocation with a provable 50\% optimality guarantee for the min-sum objective. While CBBA offers scalability, distributed decision-making, and predictable behavior, its deterministic scoring strategy often leads to globally suboptimal task assignments.

To address the suboptimality inherent in the hand-crafted scoring mechanism of CBBA, this work proposes replacing the deterministic bidding function with a learned bidding policy powered by reinforcement learning (RL) and a neural network (NN) actor. The objective is to learn a policy that achieves near-optimal performance for the min-sum objective in MRTA scenarios characterized by single-task robots, single-robot tasks, and time-extended assignment (ST-SR-TA) assumptions~\cite{MRTAReview, MRTASurvey}. By integrating policy learning directly into the auction-consensus framework, each agent computes bids using a learned strategy trained in simulation, guiding the coordination process toward solutions that more closely approximate the globally optimal solution. This approach represents a step toward unifying classical distributed coordination algorithms with modern data-driven policy learning techniques.
\section{Related Work}
\label{Related_work}

Decentralized auction-consensus methods have been widely studied for MRTA settings where agents execute multiple tasks while each task is assigned to a single agent. Among these approaches, the Consensus-Based Bundle Algorithm (CBBA) remains one of the most influential and widely adopted benchmarks~\cite{CBBA}. In CBBA, each agent greedily constructs a bundle of tasks based on marginal gain estimates derived from a locally optimized route, while a consensus phase ensures conflict resolution and convergence across the team. Extensions and related auction-consensus formulations have further explored this paradigm, such as consensus ADMM, which converges to the optimal solution for the ST-SR-IA assumption~\cite{ADMM}. Beyond auction-based methods, distributed optimization techniques have also been proposed to address the MRTA problem. A decentralized genetic algorithm (GA) was proposed and compared with CBBA and CBBA-like variants~\cite{DecGA}. The decentralized GA performed competitively with the consensus approaches for the min-max objective, but could perform worse for the min-sum objective. While these traditional approaches are effective and scalable, they rely on hand-crafted cost or scoring functions and assumptions that may limit adaptivity, which can lead to suboptimal allocations as the problem complexity increases.

Recent research has explored the use of machine learning and reinforcement learning to improve solution quality in MRTA by enabling agents to learn task allocation policies from data rather than relying on fixed heuristics. Graph-based neural networks have been proposed as a natural representation for decentralized MRTA, allowing agents to reason over variable-sized teams and task sets using local communication, as demonstrated in DGNN-GA~\cite{DGNNGA}. Other works employ deep reinforcement learning with graph encoders and normalization techniques to improve scalability and adaptability across different problem sizes~\cite{Scalable_Graph}. Capsule- and graph-based reinforcement learning methods have also been proposed to address MRTA, explicitly targeting the challenge of scaling learned policies to larger task spaces without retraining~\cite{CAM}. Additionally, decentralized reinforcement learning frameworks have been developed for heterogeneous teams, aiming to learn generalized policies applicable across different agent capabilities and task configurations~\cite{dai2025heterogeneous}. Despite these advances, many learning-based MRTA approaches remain sensitive to the training distribution, often requiring retraining or fine-tuning when the number of agents, workspace geometry, or task distribution changes. As a result, generalization across environments remains an open challenge, even though graph-based and permutation-invariant architectures have shown promising improvements.

This work contributes to the MRTA literature by bridging classical auction-consensus algorithms with modern reinforcement learning under a centralized training and decentralized execution (CTDE) framework.  Unlike prior learning-based MRTA approaches that learn allocation policies end-to-end, the proposed approach embeds a learned bidding policy within the decentralized auction-consensus framework. The combination of reinforcement learning with auction-consensus mechanisms provides a novel direction for MRTA research, offering a potential method for improving solution quality while maintaining generality across varying environments, swarm sizes, and task distributions.

\section{Auction-Consensus Algorithm}
\label{methods}

The proposed system introduces an improved auction-consensus algorithm in which the traditional CBBA scoring rule is replaced by a learned bidding policy trained using reinforcement learning. Instead of relying on handcrafted marginal gain functions, each agent employs a neural network bidder that computes bids directly from agent-specific and task-specific features. Although this modification alters the bidding mechanism, the overall structure of the algorithm remains consistent with two-phase of CBBA formulation: a task-selection phase followed by a consensus phase. 

We consider an MRTA problem with $N_u$ agents and $N_t$ tasks. Let the agent index set be $\mathcal{I} \triangleq \{1,\dots,N_u\}$ and the task index set be $\mathcal{J} \triangleq \{1,\dots,N_t\}$. Each task may be assigned to at most one agent, and each agent may execute up to $L_t$ tasks. The goal is to determine a conflict-free assignment that optimizes a global min-sum objective.

Following CBBA, each agent $i \in \mathcal{I}$ maintains four state variables: a winning bid vector $\mathbf{y}_i \in \mathbb{R}_+^{N_t}$ storing the highest known bid for each task, a winning agent vector $\mathbf{z}_i \in \mathcal{I}^{N_t}$ indicating the current believed owner of each task, a bundle $\mathbf{b}_i \in (\mathcal{J}\cup\{0\})^{L_t}$ representing the ordered set of tasks assigned to the agent, and a corresponding execution path $\mathbf{p}_i \in (\mathcal{J}\cup\{0\})^{L_t}$. These variables are updated through the auction and consensus phases to achieve decentralized conflict resolution.

Under the ST-SR-TA (Single-Task robots, Single-Robot tasks, Time-extended Assignment) classification~\cite{MRTAReview, MRTASurvey}, each agent executes at most one task at any given instant, and each task is assigned to exactly one agent. Although each agent constructs a bundle $\mathbf{b}_i$ containing up to $L_t$ tasks, these tasks are planned and executed sequentially along an ordered route $\mathbf{p}_i$, forming a time-extended assignment.

In contrast to classical CBBA, where bids are calculated using deterministic marginal gain functions, we construct a partial observation $\mathbf{o}_i(t)$ from the local state of the agent and task features. The function $\Phi(\cdot)$ constructs the partial observation from the agent's local state and the task position set $Q = \{\mathbf{q}_j \in \mathbb{R}^2 \mid j \in \mathcal{J}\}$ where $\mathbf{q}_j$ denotes the planar position of task $j$. The resulting observation is represented as $\mathbf{o}_i(t) \in \mathbb{R}^{N_t \times F}$, where $N_t$ is the number of tasks and $F$ is the feature dimension per task.
The neural bidding policy $\pi_\theta(\cdot)$ maps the partial observation to a bid vector $\mathbf{c}_i \in \mathbb{R}^{N_t}$, producing a scalar bid value per task. A higher value of $\mathbf{c}_{ij}$ is interpreted as a stronger preference to assign task $j$ to agent $i$. For tasks already contained in the agent’s bundle, the corresponding bid outputs are masked prior to task selection by setting their effective bid values
to $-\infty$ (or equivalently forcing $h_{ij}(t)=0$) for all $j \in \mathbf{b}_i(t)$. The CBBA consensus mechanism is retained to coordinate task ownership between agents, while learning replaces the handcrafted scoring function.

Phase 1 consists of constructing partial observations from current agent states, computing task bids using the learned neural policy, and greedily selecting tasks for self-assignment (see Algorithm~\ref{alg:learned_cbba_phase1}).
Each agent evaluates all available tasks by computing task-specific features, such as distances to its current position and along its current route, before generating bids. The agent compares these offers with its current winning list and selects tasks for which it can place higher bids until its bundle capacity $L_t$ is reached. The newly selected tasks are inserted into the route in the position that minimizes the increase in the path length, denoted $\Delta D(\cdot)$.

\begin{equation}
\Delta D(\mathbf{p}_i \oplus_n \{j\}) = 
D(\mathbf{p}_i \oplus_n \{j\}) - D(\mathbf{p}_i)
\end{equation}
where $D(\cdot)$ denotes the total length of the route for the agent path $i$.

\begin{algorithm}[t]
\caption{Learned Bundle Construction for Agent $i$ at Iteration $t$}
\begin{algorithmic}[1]

\Procedure{Build Bundle}{$\mathbf{z}_i(t-1), \mathbf{y}_i(t-1), \mathbf{b}_i(t-1), \mathbf{p}_i(t-1)$}
\State $\mathbf{y}_i(t) = \mathbf{y}_i(t-1)$
\State $\mathbf{z}_i(t) = \mathbf{z}_i(t-1)$
\State $\mathbf{b}_i(t) = \mathbf{b}_i(t-1)$
\State $\mathbf{p}_i(t) = \mathbf{p}_i(t-1)$

\Statex

\State Construct partial observation features:
\State $\mathbf{o}_i(t) = \Phi\!\left(\mathbf{y}_i(t), \mathbf{z}_i(t), \mathbf{b}_i(t), \mathbf{p}_i(t), Q\right)$

\Statex

\State Compute bids using learned neural policy:
\State $\mathbf{c}_i(t) = \pi_\theta\!\left(\mathbf{o}_i(t)\right)$

\Statex

\ForAll{$j \in \mathcal{J}$}
    \State $h_{ij}(t) = \mathbb{I}\!\left(c_{ij}(t) > y_{ij}(t)\right)$
\EndFor

\Statex

\While{$|\mathbf{b}_i| < L_t$}

\If{$\sum_{j \in \mathcal{J}} h_{ij}(t) = 0$}
    \State \textbf{break}
\EndIf

\State $J_i = \arg\max_{j \in \mathcal{J}} \; c_{ij}(t)\cdot h_{ij}(t)$

\State $n_{i,J_i} = \arg\min_n \; \Delta D\!\left(\mathbf{p}_i \oplus_n \{J_i\}\right)$

\State $\mathbf{b}_i = \mathbf{b}_i \oplus_{\text{end}} \{J_i\}$

\State $\mathbf{p}_i = \mathbf{p}_i \oplus_{n_{i,J_i}} \{J_i\}$

\State $y_{i,J_i}(t) = c_{i,J_i}(t)$

\State $z_{i,J_i}(t) = i$

\EndWhile

\EndProcedure
\end{algorithmic}
\label{alg:learned_cbba_phase1}
\end{algorithm}

Phase 2 consists of agents executing a consensus phase to resolve conflicts over task ownership using the standard CBBA communication update rules \cite{CBBA}. During this phase, agents exchange their current winning bid vectors $\mathbf{y}_i$ and corresponding winner identity vectors $\mathbf{z}_i$ for all tasks. Based on the received information, each agent updates its local variables $\mathbf{y}_i$, $\mathbf{z}_i$, $\mathbf{b}_i$, and $\mathbf{p}_i$ according to the CBBA update, reset, or leave rules.

Unlike classical CBBA, no diminishing marginal gain (DMG) condition is enforced on the learned bidding policy. Therefore, the standard convergence and optimality guarantees of CBBA do not necessarily apply. The consensus mechanism is retained to promote consistency of task assignments across agents under decentralized communication.

\section{Reinforcement Learning Design and Training}
\label{Reinforcement Learning}

The neural bidding model is trained using a Centralized Training with Decentralized Execution (CTDE) framework, which is well-suited for multi-agent cooperation. During training, all information required to evaluate global team performance, including the optimal assignment and optimal team distance computed via a MILP solver~\cite{PULP}, is available to the learning algorithm. This allows the system to shape rewards based on how closely the team’s collective decisions approximate the optimal task allocation solution. To enable decentralized execution, each agent’s observation, both during training and testing, is only computed from each agent's locally available state information. These partial observations form the sole input to the actor network, ensuring that the learned bidding policy does not rely on global knowledge. In contrast, the critic network receives a global observation summarizing the overall system state, which is used exclusively during training to stabilize learning and encourage cooperative behavior among agents.
CTDE, therefore, allows agents to learn cooperative behaviors during training while still respecting the distributed constraints inherent to auction-consensus execution at runtime. The reinforcement learning paradigm used for this work is Proximal Policy Optimization (PPO), which collects rollouts from many randomly generated MRTA worlds. These rollouts capture the interactions between agents, their evolving bundles and paths, and the auction-consensus mechanics that determine how bidding decisions propagate through the team. Because the critic network has access to global information while the actor does not, training encourages agents to implicitly coordinate through their bids and learn a policy that helps the group approach the optimal solution without ever requiring centralized communication at execution time.

The reward function measures improvements in collective performance by combining normalized changes in team travel distance, task assignment coverage, and agreement with the optimal assignment. 

The reward at iteration $t$ is defined as
\begin{equation}
r(t) = w_D r_D(t) + w_C r_C(t) + w_A r_A(t)
\end{equation}
where $r_D$ measures normalized improvement in team travel distance,
$r_C$ measures normalized improvement in task coverage, and $r_A$ measures normalized improvement in agreement with the MILP optimal assignment. The $w$ variables denote the corresponding weights in the final reward weighted sum.

Each reward component is scaled to be within the range [-1,1] and combined using a weighted sum to guide learning toward higher-quality global allocations without directly enforcing optimality.

\begin{figure}[h]
  \centering
  \includegraphics[width=0.45\textwidth]{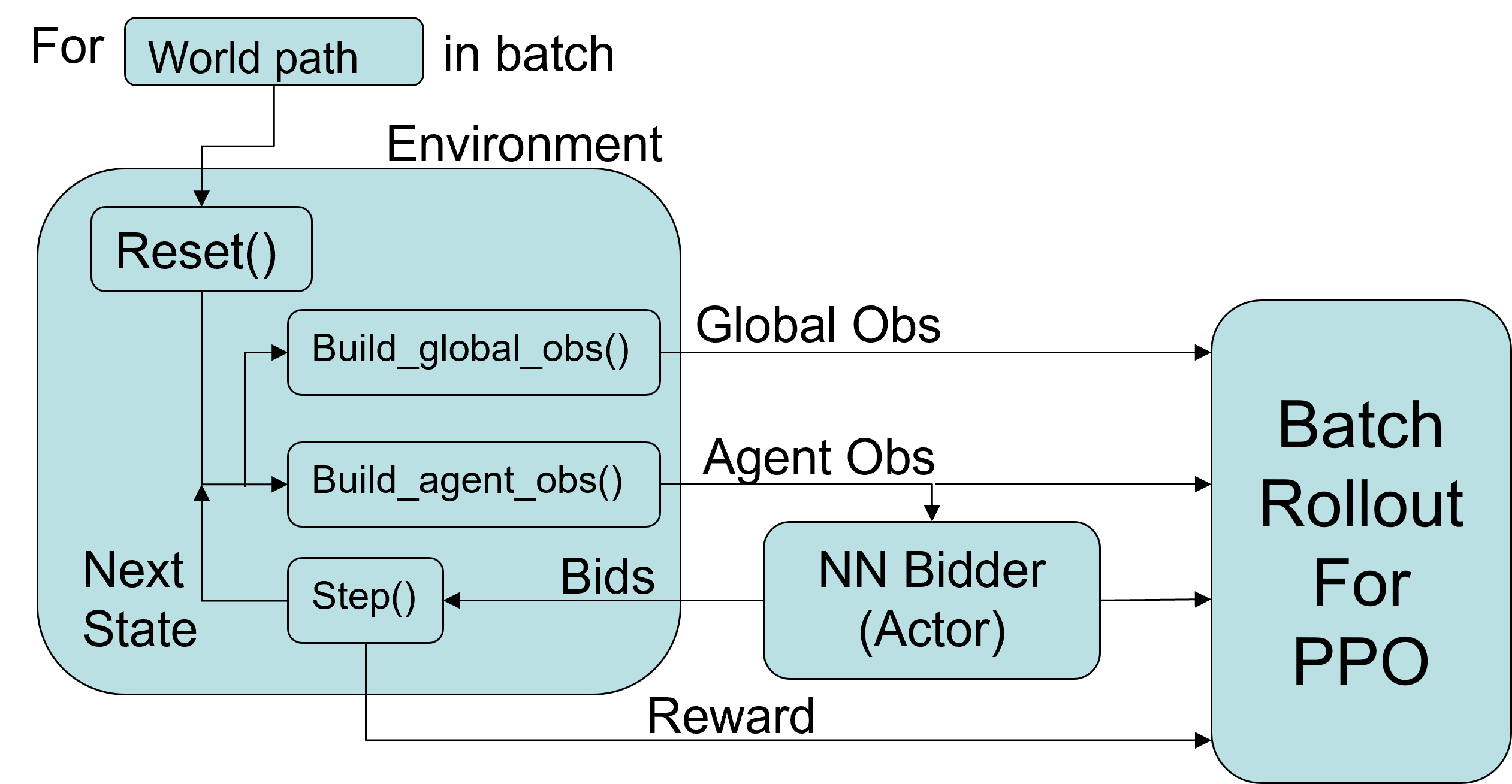}
  \caption{Training Pipeline}
  \label{Training_Pipeline}
  
\end{figure}

Training is conducted over multiple epochs, each consisting of a batch of randomly generated MRTA worlds. For each world, agents apply the bidding policy being trained until convergence or timeout, producing trajectories of observations, actions (bids), and rewards that are used to update the actor and critic networks via PPO as shown in Fig. \ref{Training_Pipeline}. In Fig. \ref{Training_Pipeline}, the block "Build\_agent\_obs$()$" corresponds to lines 2-7 and the block "NN Bidder $($Actor$)$" corresponds to lines 8-9 from Alg. \ref{alg:learned_cbba_phase1}. The block "Step$()$" represents the rest of the Phase 1, Phase 2, and is also where the reward is calculated.   After training, the actor, which corresponds to the learned bidding policy, is evaluated on a separate validation set that includes larger swarms and expanded workspaces, allowing the generality and scalability of the decentralized execution policy to be assessed.

All learned bidding policies were trained using a dataset of 1000 randomly generated 2D MRTA worlds, each containing 5 agents and a variable number of tasks ranging from 10 to 20. The tasks and agent positions were uniformly sampled within square workspaces whose side lengths varied from $25\times25$ to $55\times 55$, introducing significant geometric diversity during training.

Three neural bidding architectures were evaluated: Neural Additive Model (NAM), Long Short-Term Memory (LSTM)-Based Model, and Set Transformer-Based Model.

\begin{figure}[h]
  \centering
  \includegraphics[width=0.35\textwidth]{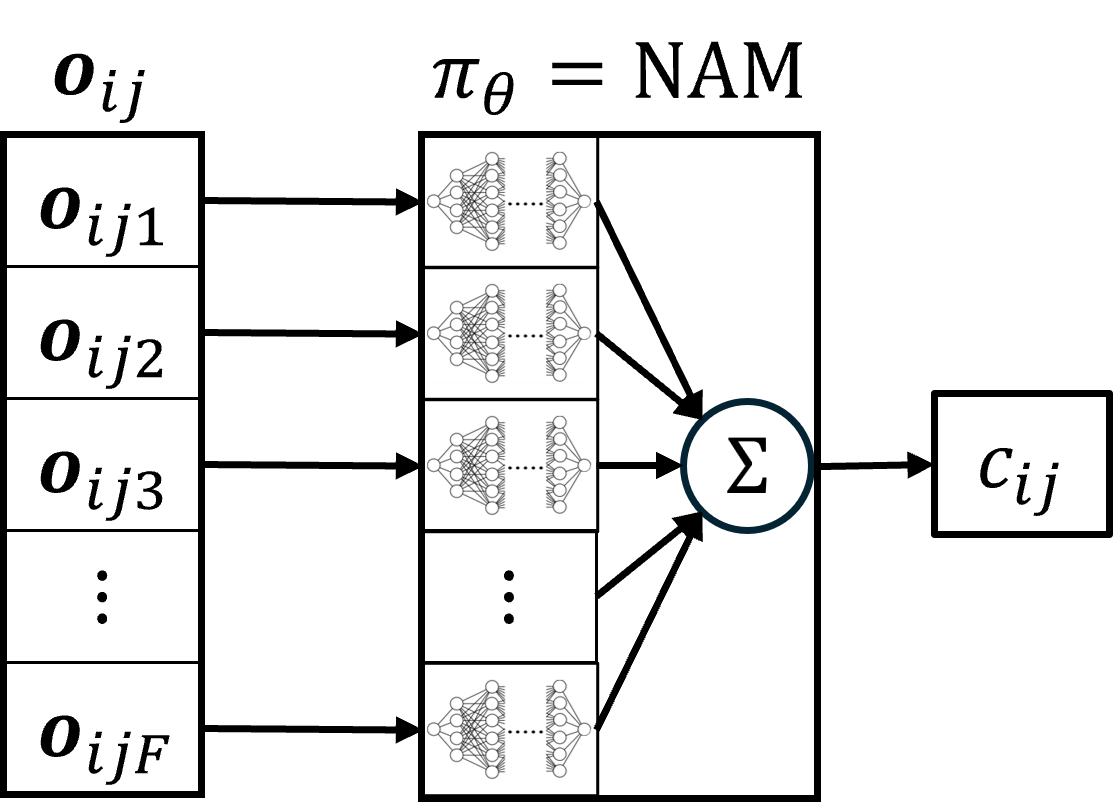}
  \caption{Neural Additive Model as the Neural Bidder}
  \label{NAM_Diagram}
  
\end{figure}

NAM was included as a lightweight and interpretable baseline for the neural network ~\cite{NAM_ref}. Its additive structure, as shown in Fig. \ref{NAM_Diagram}, allows direct inspection of how individual task features contribute to bid values. Although the NAM achieved strong average performance, it exhibited convergence failures (timeouts), where auction-consensus iterations failed to reach a consistent allocation solution within the maximum iteration limit. When evaluated in the validation datasets described in Section V, this occurred in a subset of validation worlds, with 95 timeouts for 5-agent swarms and 7 timeouts for 20-agent swarms out of 1000 worlds each.

\begin{figure}[h!]
  \centering
  \includegraphics[width=0.35\textwidth]{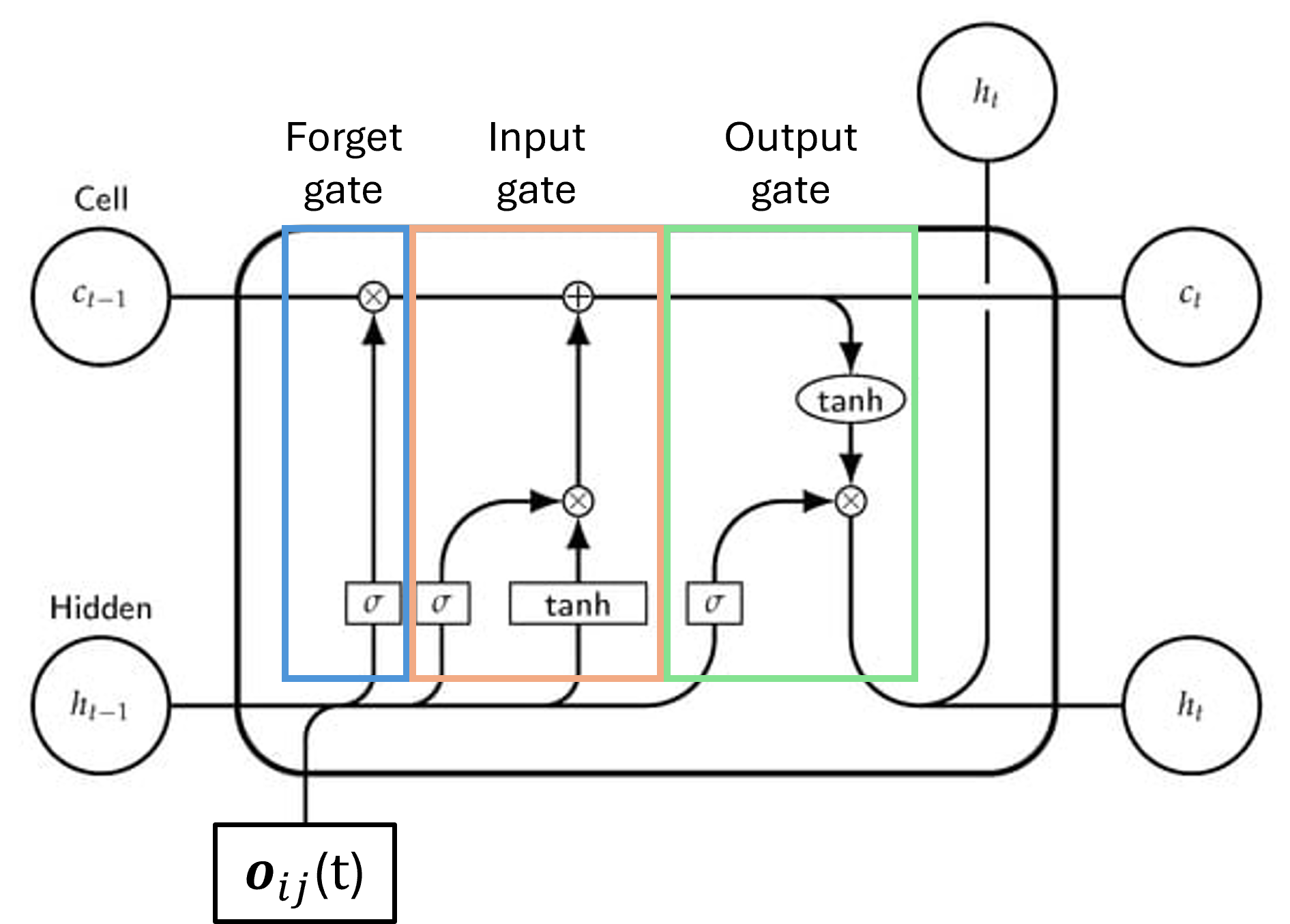}
  \caption{LSTM Cell Structure}
  \label{LSTM_Cell}
      \vspace{0.05in}
\end{figure}

\begin{figure}[h!]
  \centering
  \includegraphics[width=0.35\textwidth]{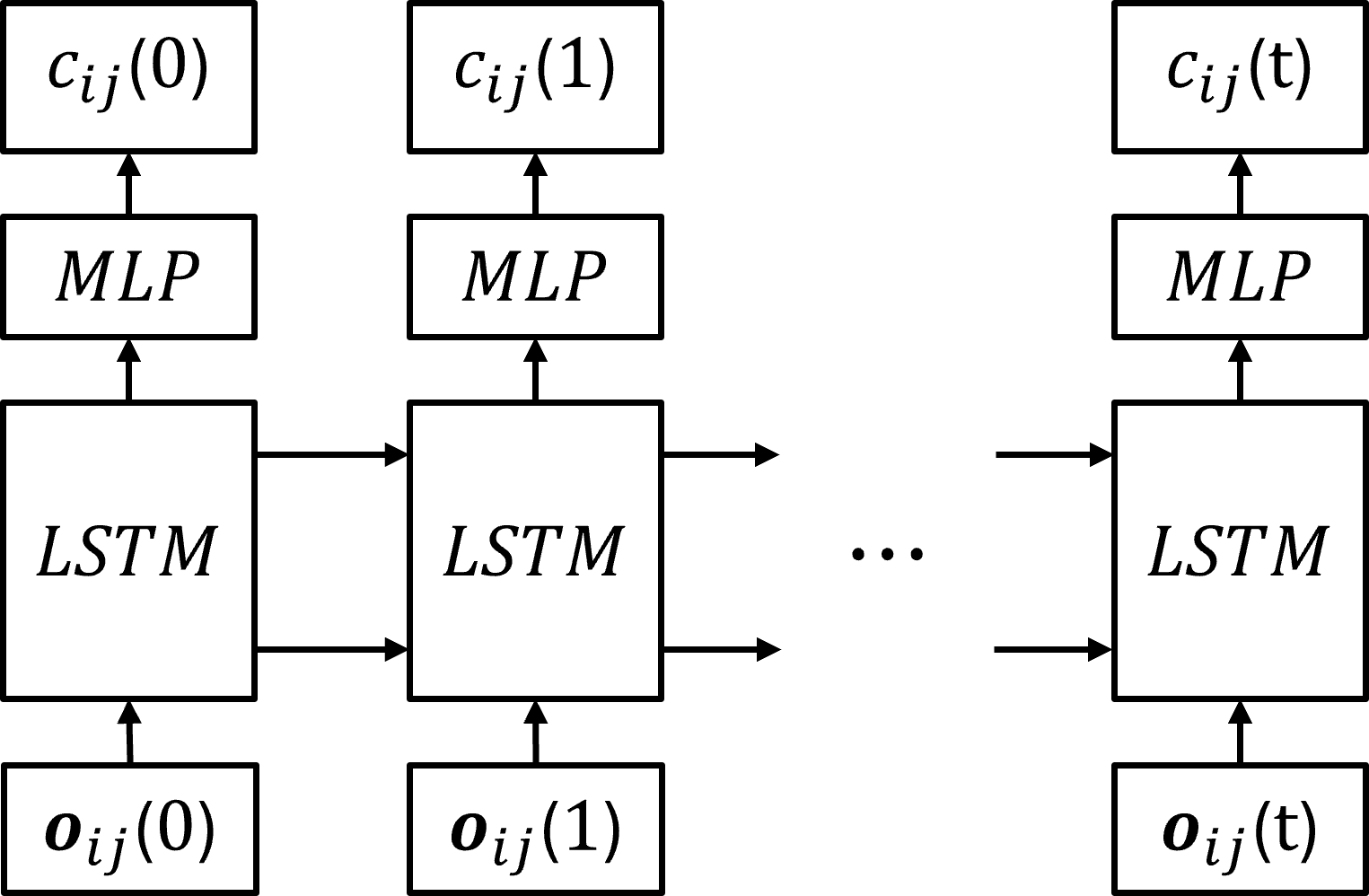}
  \caption{LSTM-Based Model as the Neural Bidder}
  \label{LSTM_Diagram}
      \vspace{-.15in}
\end{figure}

\begin{figure}[h!]
  \centering
  \includegraphics[width=0.5\textwidth]{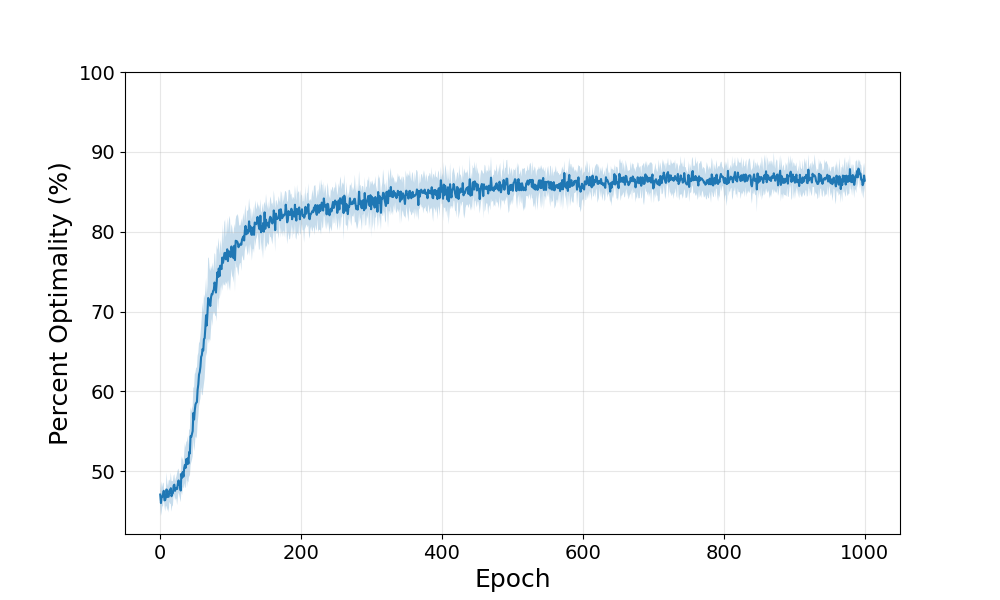}
  \caption{Mean training curve for the LSTM actor across 10 training runs}
  \label{LSTM_Training_Curve}
\end{figure}

The LSTM bidder was evaluated due to its ability to maintain memory, which is well-suited for decentralized MRTA framed as a partially observable decentralized  Markov decision process (Dec-POMDP)~\cite{POMDP_Recurrent}. An LSTM cell stores a cell state and a hidden state, which are updated as new inputs are given, as shown in Fig.~\ref{LSTM_Cell}. This enables agents to possess memory and form an implicit belief state from a history of partial observations. Fig. \ref{LSTM_Diagram} shows how, at a given timestep, $\mathbf{c}_{ij}(t)$ is calculated using observations from that timestep and hidden features passed along from previous observations. The hidden state at a given timestep serves as an encoded belief state that is fed into a multi-layer perceptron (MLP) head to compute the bid value for that timestep. Since most reinforcement learning algorithms are designed for fully observable MDPs, learning in belief space substantially improves policy effectiveness. Training results strongly support this hypothesis; Fig. \ref{LSTM_Training_Curve} shows how the mean training curve averaged over 10 independent LSTM-based runs consistently converged to approximately 87\% optimality, demonstrating both stability and repeatability.

\begin{figure}[h!]
  \centering
  \includegraphics[width=0.5\textwidth]{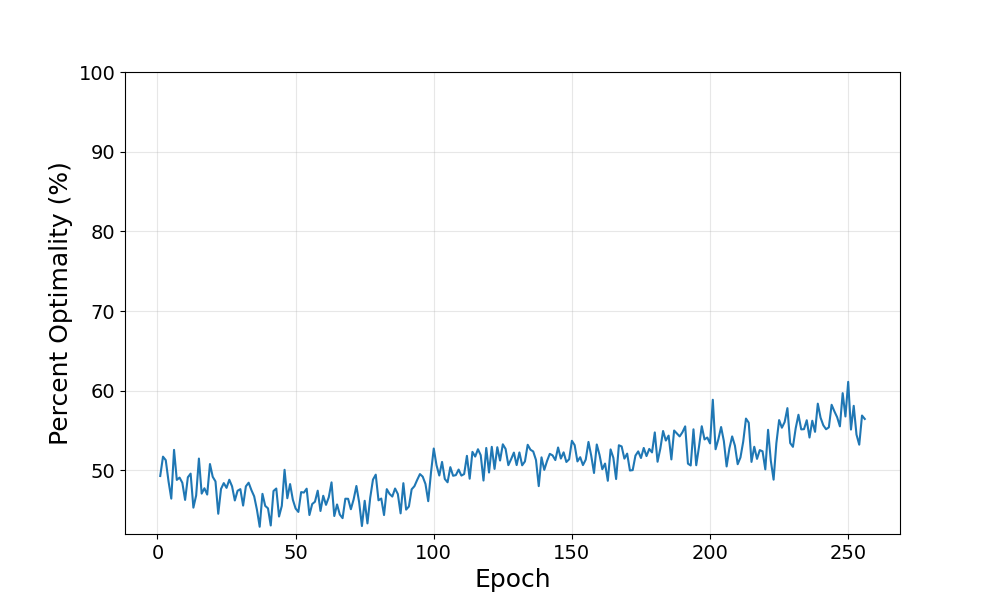}
  \caption{Set Transformer actor training curve}
  \label{Transformer_Training_Curve}
  
\end{figure}

A set transformer model was tested due to the appearance of attention-based architectures in recent MRTA and vehicle routing literature~\cite{AMARL, SADCHER}. However, the transformer exhibited unstable training dynamics in its training curve, as seen in Fig. \ref{Transformer_Training_Curve}, and failed to learn an effective policy. 

\begin{figure}[h!]
  \centering
  \includegraphics[width=0.5\textwidth]{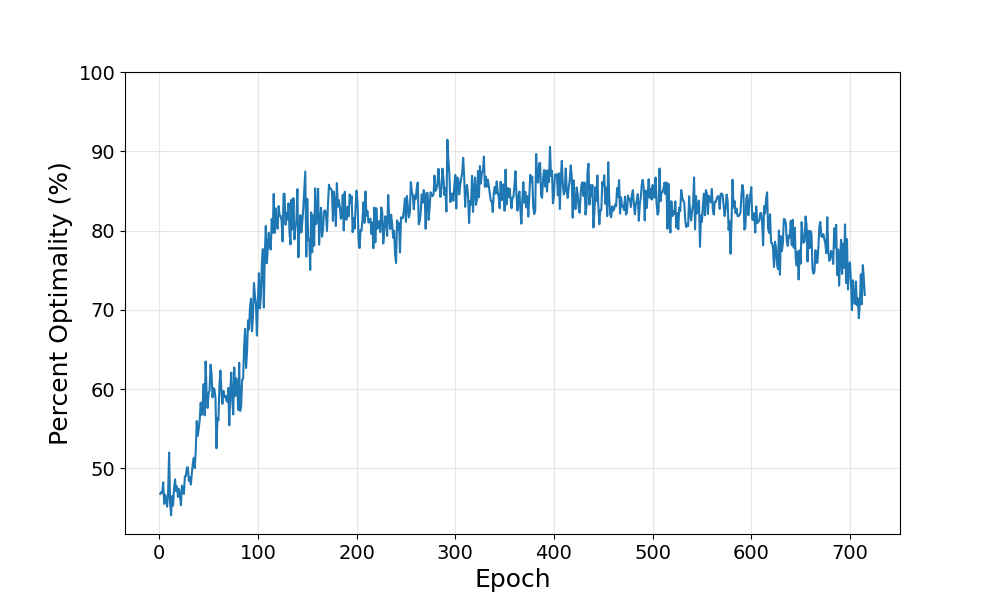}
  \caption{Hybrid Set Transformer and LSTM actor training curve}
  \label{Hybrid_Training_Curve}
    \vspace{-.15in}
\end{figure}

A hybrid architecture, consisting of a set transformer encoder followed by an LSTM, was also evaluated to inject global context into the recurrent model. Although this hybrid approach initially reached approximately 87\% optimality, the hybrid training curve in Fig. \ref{Hybrid_Training_Curve} shows the performance later degraded to roughly 73\%, indicating instability and poor convergence behavior as well. As a result, both transformer-based variants were excluded from final comparative evaluation. 

\section{Experimental results}
\label{results}


Final performance comparisons were conducted among CBBA, the NAM actor, and the LSTM actor, using percent optimality and the number of iterations to convergence as the main metrics.
Percent optimality is formally defined as
\begin{equation}
    \eta = \frac{D^*}{\hat{D}} \times 100\%,
    \label{eq:percent_optimality}
\end{equation}

where $D^*$ denotes the total distance traveled by the team of the optimal assignment obtained through the MILP solver, and $\hat{D}$ denotes the total distance traveled by the team of the final assignment produced by the evaluated algorithm upon convergence. A value of $\eta = 100\%$ indicates that the algorithm matches the optimal solution exactly, while lower values reflect increasing suboptimality.

To evaluate generalization, trained models were tested on separate validation datasets, each consisting of 1000 unseen worlds for swarm sizes of 5, 10, 15, and 20 agents. The task counts were scaled proportionally to the size of the swarm, ranging from 2 to 4 tasks per agent, resulting in 10–20 tasks for 5-agent swarms and 40–80 tasks for 20-agent swarms. All validation worlds followed the same workspace size distribution as training, but were independently sampled.


Excluding timeout cases, the NAM actor achieved the highest median optimality across all swarm sizes, reaching approximately 90\% optimality for 5-agent swarms and gradually decreasing to a median of around 87\% for 20-agent swarms. These results indicate that the learned additive scoring can outperform the CBBA hand-crafted scoring rule when convergence is achieved. With further work to mitigate timeouts, the NAM actor represents a strong alternative to classical CBBA scoring.

The LSTM actor also outperformed CBBA in smaller and medium swarm sizes with a median optimality roughly $5\%$ better than 5-agent swarms. The LSTM actor also did not exhibit convergence failures in all validation sets. However, the performance of the LSTM actor degraded with increasing swarm size, and by 20 agents, the median optimality of the LSTM actor was almost equal to that of CBBA. This suggests that while memory-based bidding mitigates partial observability effects, additional architectural or coordination mechanisms may be required to maintain performance at larger scales.

\begin{figure}[h!]
  \centering
  \includegraphics[width=0.48\textwidth]{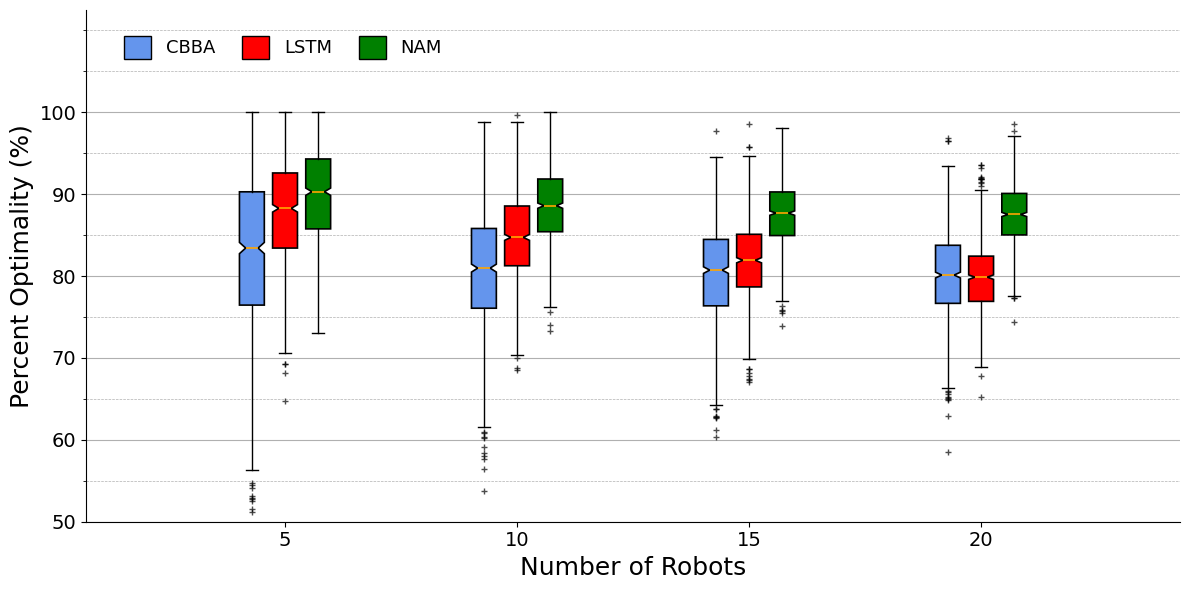}
  \caption{Percent Optimality vs Swarm Size}
  \label{Optimality_Comparison}
\end{figure}

The convergence speed was evaluated by measuring the number of auction-consensus iterations required to reach a final allocation. Across all swarm sizes, both learned bidders exhibited slower convergence compared to CBBA, whose upper-tail convergence remained stable at approximately 6--9 iterations across all swarm sizes tested. The NAM actor required more iterations on average and displayed substantially heavier tails in the iteration distribution, 
with upper-tail convergence reaching up to 21 iterations for 20-agent swarms, reflecting their convergence instability. The LSTM actor showed better convergence stability than the NAM actor, with upper-tail convergence between 11 and 13 iterations across all swarm sizes, although still noticeably slower than CBBA.

\begin{figure}[h!]
  \centering
  \includegraphics[width=0.48\textwidth]{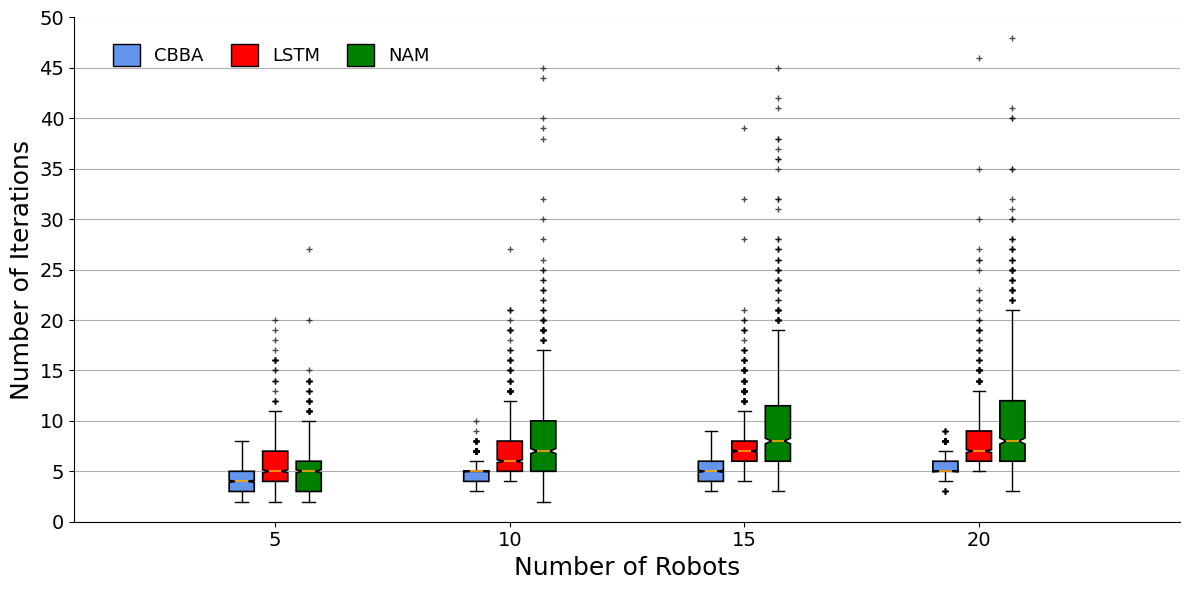}
  \caption{Number of Iterations vs Swarm Size}
  \label{Iterations_Comparison}
\end{figure}

Overall, the results demonstrate that learned bidding policies can improve the quality of the solution over CBBA while preserving decentralized execution and convergence properties. The LSTM model offers the best balance between performance, stability, and reliability, especially at smaller swarm sizes, whereas the NAM model achieves the highest optimality when convergence is successful. Transformer-based approaches, while theoretically attractive, proved ineffective in this decentralized auction-consensus setting.
\section{Conclusion}
\label{Conclusion}

This work demonstrates the feasibility and effectiveness of integrating reinforcement learning into decentralized auction-consensus algorithms for multi-robot task allocation. By replacing the fixed greedy scoring function of CBBA with a learned bidding policy trained under a centralized training and decentralized execution framework, agents can produce task allocations that more closely approach optimal solutions while preserving decentralized execution and consensus-based coordination.

Experimental results show improvements in solution quality, although with weaker convergence speed, along with encouraging signs of generalization to larger and more complex environments. The proposed approach highlights the potential of combining classical distributed coordination methods with modern learning-based techniques, enabling decentralized multi-robot systems to exhibit cooperative behaviors that were previously difficult to achieve using hand-crafted heuristics alone. This work lays the foundation for further exploration of learning-enhanced auction-consensus algorithms and their application to increasingly complex multi-robot systems.

Several avenues exist to further improve and extend the proposed framework. An important direction involves refining the auction-consensus rules to improve stability and convergence when combined with learned bidding policies, particularly in larger and more densely populated task environments. Another key area of development is the expansion of the task-feature space to incorporate additional contextual information, such as task rewards, loss-based mechanisms, or measures of inter-task proximity, which may enable the neural bidder to reason more effectively about long-term team performance.
Future work will include more extensive evaluations across a broader range of swarm sizes, task distributions, and workspace geometries to fully characterize the generality and robustness of the learned bidding policy. These efforts will help determine whether the approach can consistently outperform CBBA in diverse MRTA scenarios.






\section*{ACKNOWLEDGMENT}

The authors would like to acknowledge funding provided by the NSF (National Science Foundation) CREST Center for Multidisciplinary Research Excellence in Cyber-Physical Infrastructure Systems (No. 2112650), the NSF CISE Minority Serving Institute (MSI) program (No. 2318682 \& No. 2434916), and NSF Expand AI program (No. 2434916).




\bibliographystyle{IEEEtran}
\bibliography{BibFiles/IEEEabrv, BibFiles/main_ref}

\end{document}